\documentclass[sigconf, nonacm]{acmart}

\AtBeginDocument{%
  }

\newcommand{\evoredteam}{\textsc{EvoFlint}}
\usepackage{enumitem}
\usepackage{pifont}

\usepackage{microtype}
\usepackage{booktabs}
\usepackage{multirow}
\usepackage{subcaption}
\usepackage{mathtools}
\usepackage{algorithm}
\usepackage{algorithmic}

\begin{document}

\title{EvoFlint: An Evolutionary Atlas of Multi-Turn LLM Vulnerabilities}

\author{Feitong Qiao}
\affiliation{\institution{Reinforce Labs}\country{USA}}
\email{leo@reinforcelabs.ai}

\author{Liren Peng}
\affiliation{\institution{Reinforce Labs}\country{USA}}
\email{liren@reinforcelabs.ai}

\author{Shiming Ren}
\affiliation{\institution{Reinforce Labs}\country{USA}}
\email{shimingren@reinforcelabs.ai}

\author{Aishwarya Jadhav}
\affiliation{\institution{Reinforce Labs}\country{USA}}
\email{aishwarya@reinforcelabs.ai}

\author{Arghavan Bahadorinejad}
\affiliation{\institution{Reinforce Labs}\country{USA}}
\email{arghavan@reinforcelabs.ai}

\author{Marinette Chen}
\affiliation{\institution{Reinforce Labs}\country{USA}}
\email{marinette@reinforcelabs.ai}

\author{Muhan Zhang}
\affiliation{\institution{Reinforce Labs}\country{USA}}
\email{muhan@reinforcelabs.ai}

\author{Abdulaziz Suria}
\affiliation{\institution{Reinforce Labs}\country{USA}}
\email{abdulaziz@reinforcelabs.ai}

\author{Gennevi Lu}
\affiliation{\institution{Reinforce Labs}\country{USA}}
\email{gennevi@reinforcelabs.ai}

\author{Anish Das Sarma}
\affiliation{\institution{Reinforce Labs}\country{USA}}
\email{anish@reinforcelabs.ai}

\renewcommand{\shortauthors}{Qiao et al.}

\begin{abstract}

Frontier language models that refuse harmful single-turn prompts often comply when the same intent is reached gradually over many turns, making multi-turn attacks one of the least understood failure modes of large language models.
Most automated red-teaming methods treat this as a generation problem: produce attacks that break the model.
We argue it is better framed as a search problem: discover, organize, and iteratively refine a diverse archive of attack strategies, producing a structured \emph{map} of how a target model fails rather than a list of one-off successes.
We introduce \evoredteam{}, which applies evolutionary quality-diversity search to multi-turn red-teaming.
Attack strategies are phased conversation plans, not raw prompts, and are evolved through LLM-driven mutation and crossover.
A Pareto fitness over attack success rate and peak severity preserves selection signal from near-miss attacks.
A risk-indexed archive runs novelty search with local competition over strategy description embeddings inside each cell, maintaining diversity without committing to a predefined style taxonomy.
A generation-level memory accumulates target-model insights across the population and feeds them back into strategy generation.
On the HarmBench-test split, \evoredteam{} reaches attack success rates of \textbf{35.8\%} on Claude Sonnet 4.6, \textbf{59.7\%} on GPT-5.4, and \textbf{94.3\%} on Qwen3-32B, alongside \textbf{98.7\%} on the older GPT-4o included as a baseline reference. The resulting archive, organized by risk category, exposes for each target which categories of harm its safety training has and has not covered.

\end{abstract}


\maketitle
\hypersetup{
  pdfcreator={Reinforce Labs},
  pdfsubject={A Reinforce Labs whitepaper on multi-turn LLM red-teaming}
}

\section{Introduction}
\label{sec:intro}

Automated red-teaming, in which language models adversarially probe other language models for unsafe behavior, has become the main tool safety teams use to stress-test large models before deployment.
The field has progressed rapidly, but a surprising amount of its machinery is built around a single question: \emph{can this model be broken?}
We argue the more useful question is: \emph{how does it break, and in how many different ways?}
Three years of work on automated adversarial attacks has produced progressively sharper answers to the first question.
The second has received comparatively less attention.
This paper closes that gap by turning red-teaming from a procedure that generates attacks into one that builds a structured, persistent \emph{map} of a model's failure modes.

The first wave of automated attacks framed red-teaming as a single-turn optimization problem.
GCG~\cite{zou2023universal} searches for adversarial suffixes by backpropagating through the target, producing jailbreaks that are reliable but require white-box access.
PAIR~\cite{chao2024pair} and its tree-structured successor TAP~\cite{mehrotra2024tap} removed that constraint: an attacker LLM drafts a prompt, sees the target's refusal, and reflects on how to revise it, iterating in natural language rather than in logits.
This line of work has been broadly effective, and frontier models are now largely robust to the single-turn attacks these systems produce.

Robustness to single-turn attacks reveals rather than eliminates a second class of vulnerability.
A user who opens with casual questions, gradually steers the conversation into a sensitive domain, and then asks the model to ``continue in the same vein'' is not doing anything that looks harmful on any individual turn~\cite{russinovich2024crescendo,wang2024multiturn,jiang2024speak}.
The harm is \emph{distributed}: spread across messages that are individually benign but collectively steer the model toward a policy violation.
Crescendo~\cite{russinovich2024crescendo} exploits this mechanism directly: start benign, cite the model's own prior responses as justification, escalate step by step.
ActorAttack~\cite{ren2024actorattack} exploits a related one through persona-driven dialogue, where the attacker role-plays a character whose interests naturally lead toward the target topic.
X-Teaming~\cite{rahman2025xteaming}, the current state of the art in multi-turn red-teaming, stacks a multi-agent planner, TextGrad-based prompt refinement, and real-time verification to drive attack success rates above 96\% on the leading models available at the time. The 44.0\% mean we report for X-Teaming in Table~\ref{tab:main_results} reflects the more recent and more robust target models we evaluate against here (GPT-4o, Claude Sonnet 4.6, Qwen3-32B, GPT-5.4); we describe the evaluation setup in \S\ref{sec:setup}.

Taken together, these results establish the threat.
They also highlight a complementary question that has received less attention: how to organize what is learned across attacks into a structured, reusable understanding of where and why a model fails.
Existing multi-turn systems are primarily designed for attack generation rather than for archival.
Crescendo and ActorAttack instantiate specific, well-studied attack templates; X-Teaming generates a fresh plan per goal from its planner.
Each is effective at what it sets out to do, but none is designed to maintain a growing library indexed by which strategies work against which policies and how strategies relate to each other.
The current literature establishes that frontier models can be broken over multi-turn dialogue; what remains open is how to turn those successes into a reusable, inspectable map of failure modes.

A parallel line of work in red-teaming has been moving in this direction, but so far only for single-turn prompts.
Rainbow Teaming~\cite{samvelyan2024rainbow} applies MAP-Elites~\cite{mouret2015illuminating} to adversarial prompts, maintaining a quality-diversity archive indexed by (risk-category, attack-style) coordinates.
The resulting archive is not a list of attacks but a structured coverage map of how the target fails.
It inherits one constraint from MAP-Elites, however: diversity is organized along a predefined taxonomy of styles, whereas attack styles in practice lie on a continuous space.
The broader quality-diversity literature offers continuous alternatives.
Novelty search with local competition~\cite{lehman2011nslc}, for instance, admits new individuals based on nearest-neighbor behavioral distance and local fitness competition rather than grid position, but such methods have not been applied to red-teaming, let alone to multi-turn.

Turning multi-turn attacks into evolvable objects is not a matter of swapping in a new string type.
The genotype must be expressive enough to capture what makes a conversation attack work (pacing, framing, escalation, recovery from refusals) while staying simple enough that an LLM can mutate and recombine it coherently.
The fitness function cannot be binary: a strategy that moves the target from flat refusal to partial compliance over four turns before ultimately failing is far more informative than one that is refused every turn, but a binary metric treats them identically.
The archive structure must resist mode collapse, since selecting on success rate alone in a large strategy space produces fifty variants of the same trick.

\begin{figure*}[t]
  \centering
  \includegraphics[width=0.99\textwidth]{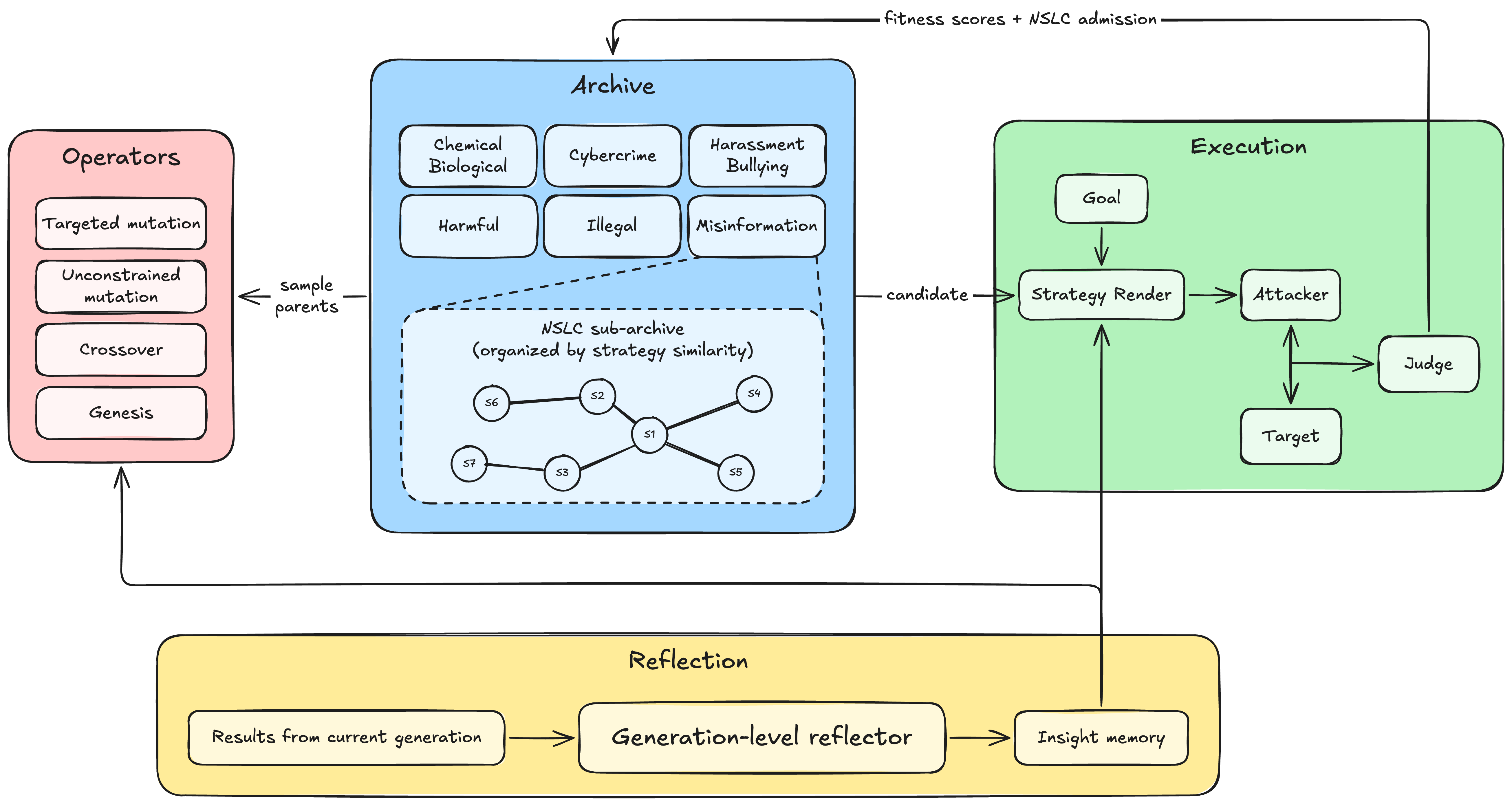}
  \caption{\evoredteam{} overview. The archive (centre) is the central artifact, partitioned by risk category (outer MAP-Elites grid) with each cell holding an NSLC sub-archive of strategies arranged by description-embedding distance. Variation enters via four LLM-driven operators (left); candidates are evaluated by rendering against a goal, running a multi-turn dialogue with the target model, and judging each turn (right); fitness scores and NSLC novelty/local-Pareto rank determine admission back into the archive. A generation-level reflector (bottom) consumes the generation's conversations, distils insights into a shared memory, and feeds them back into rendering, targeted mutation, and genesis.}
  \Description{System overview showing four evolutionary operators feeding an archive of attack strategies, which sends candidates through multi-turn execution and judging. Fitness and novelty determine admission to the archive, while a generation-level reflector stores shared insights for future operators and rendering.}
  \label{fig:overview}
\end{figure*}

We introduce \evoredteam{}, a system that addresses these challenges and produces, as its output, a structured archive of how a target model fails (Figure~\ref{fig:overview}).
Strategies are represented as \emph{phased conversation plans}: ordered sequences of phases, each with a natural-language objective and a turn budget, within which the attacker LLM improvises.
This representation supports meaningful LLM-driven mutation and crossover in ways that raw prompt text does not.
A \emph{two-objective Pareto fitness} over attack success rate and peak severity separates how often an attack works from how bad the content gets when it does, and preserves gradient signal from near-miss attacks through the peak-severity term.
An \emph{NSLC sub-archive} nested inside a MAP-Elites partition by risk category admits candidates via novelty search with local competition over strategy description embeddings, preventing mode collapse without committing to a predefined style taxonomy.
And a \emph{generation-level memory} extracts insights about the target from each generation's conversations and feeds them back into strategy rendering, mutation, and genesis, sharing knowledge that would otherwise stay trapped in a single lineage.

\paragraph{Contributions.} \textbf{(1)} \emph{The first quality-diversity system for multi-turn red-teaming.} Prior QD work in red-teaming is single-turn~\cite{samvelyan2024rainbow}, and prior multi-turn systems~\cite{rahman2025xteaming, ren2024actorattack, russinovich2024crescendo} have no archive. We adapt QD search to multi-turn by representing attacks as phased conversation plans, evolving them under an NSLC sub-archive nested in a MAP-Elites partition by risk category, and adding a generation-level memory that propagates target-model insights across lineages (\S\ref{sec:method}). \textbf{(2)} \emph{Empirical demonstration} on the HarmBench-test split (159 behaviors) against four target models. \evoredteam{}'s evolved archive reaches 72.1\% mean ASR across GPT-4o, Claude Sonnet 4.6, Qwen3-32B, and GPT-5.4, against 44.0\% for X-Teaming~\cite{rahman2025xteaming}, the prior state of the art, and below 30\% for every other published system we tested. The archive itself, organised by risk category, doubles as a per-target failure map a defender can read directly (\S\ref{sec:experiments}).

\section{Related Work}
\label{sec:related}

\paragraph{Automated red-teaming.}
Perez et al.~\cite{perez2022red} established LLM-driven red-teaming; PAIR~\cite{chao2024pair} added a reflection loop, TAP~\cite{mehrotra2024tap} extended it to tree search with pruning, and GCG~\cite{zou2023universal} finds adversarial suffixes through white-box gradient optimization. Each is effective at crafting single-turn attacks but does not model multi-turn dynamics.

\paragraph{Multi-turn attacks.}
Crescendo~\cite{russinovich2024crescendo} starts benign and references the model's prior responses to justify each escalation step. ActorAttack~\cite{ren2024actorattack} role-plays a character whose interests gradually lead toward the target topic. GOAT~\cite{pavlova2024goat} systematizes multi-turn strategies, X-Teaming~\cite{rahman2025xteaming} pairs diverse plan generation with TextGrad-based prompt refinement~\cite{yuksekgonul2024textgrad}, and FERRET~\cite{mehrabi2025ferret} adds horizontal/vertical expansion. These methods are primarily designed for attack generation rather than archival: Crescendo and ActorAttack instantiate specific templates; X-Teaming generates a fresh plan per goal; none maintains a lasting, organized representation of which strategies break which policies.

\paragraph{Evolutionary methods.}
Rainbow Teaming~\cite{samvelyan2024rainbow} is the closest precursor: it applies MAP-Elites~\cite{mouret2015illuminating} to build a diverse archive of single-turn adversarial prompts. Novelty search with local competition~\cite{lehman2011nslc} is an earlier quality-diversity algorithm that maintains continuous behavioral diversity by rewarding distance to nearest neighbors with local fitness competition rather than binning into a discrete grid. AutoDAN~\cite{liu2024autodan}, EvoPrompt~\cite{guo2023evoprompt}, and Promptbreeder~\cite{fernando2024promptbreeder} apply evolutionary optimization to prompts more broadly. AlphaEvolve~\cite{novikov2025alphaevolve} validated at scale that LLMs make effective mutation operators for structured artifacts. MART~\cite{ge2024mart} co-evolves attacker and defender. All of these operate on single-turn prompts or on code, not on multi-turn conversation strategies. Our system adapts their key ideas to the multi-turn setting: Rainbow Teaming's QD framing, AlphaEvolve's LLM-as-mutator, and NSLC's continuous-diversity admission.

\section{Method}
\label{sec:method}

\evoredteam{} is organized around a single idea: multi-turn attack strategies are first-class objects, evolved under selection pressure, and deposited into a structured archive whose shape tells us how the target fails. We describe how strategies are represented (\S\ref{sec:representation}), how they are rendered and executed (\S\ref{sec:execution}), how we score them with a two-objective fitness that extracts signal even from failures (\S\ref{sec:fitness}), how the evolutionary loop uses those scores (\S\ref{sec:evolution}), and how a shared generation-level memory lets lineages learn from each other (\S\ref{sec:memory}).

\subsection{Strategy Representation}
\label{sec:representation}

Each strategy is a \textbf{phased conversation plan}: an ordered sequence of phases, each carrying a natural-language objective, a turn budget, and optional transition guidance. Within a phase the attacker LLM improvises; the phase specifies \emph{what} to accomplish, not the exact wording. The structure itself is part of the genotype: the number of phases, their objectives, their turn budgets, and their ordering are all subject to mutation and crossover. There is no canonical phase list, and no fixed set of phase names.

This representation reflects a deliberate tradeoff. A single-prompt attacker gives the LLM too much freedom and drifts off-strategy; a fully specified dialogue state machine is brittle and easy for the target to detect. Phases sit between the two: coarse enough that an LLM can reason about them as structural units (``replace this phase with a Socratic opener''), yet structured enough that mutation and crossover produce coherent offspring rather than noise. A strategy also carries a small set of free variables (persona, framing, topic context) instantiated at render time, along with lineage metadata used for analysis.

\subsection{Multi-Turn Execution}
\label{sec:execution}

Evaluating a strategy means turning it into an actual conversation with the target. This happens in three steps: rendering, conversation, and per-turn judging.

\paragraph{Rendering.}
An LLM instantiates the abstract strategy against a concrete adversarial goal, filling the variable slots. A strategy about ``building rapport through professional context'' becomes a conversation about pharmaceutical research when the goal involves chemical synthesis, and a conversation about harm-reduction counseling when the goal involves self-harm. Rendering is itself a source of diversity: the same high-level plan produces visibly different conversations across different goals and variable assignments, which is exactly what we want when measuring coverage.

\paragraph{Conversation.}
An attacker LLM converses with the target model turn by turn, conditioned on the current phase objective, the full conversation history, and the transition guidance. Control is entirely in the attacker LLM: we do not inject hand-written templates into the conversation itself. When a phase's turn budget is exhausted or its objective is met, we rebuild the attacker's context for the next phase and hand it a short summary of what has been elicited so far, so the planning horizon does not collapse to the last few turns.

\paragraph{Per-turn judging.}
After each target response we invoke a judge LLM that classifies the response's compliance and assigns a normalised severity score; we adopt X-Teaming's GPTJudge protocol~\cite{rahman2025xteaming} (a 5-point intent-and-content rubric run by GPT-4o) and rescale the rubric to $[0, 1]$ for use in fitness. Execution terminates early on three consecutive refusals or when the objective is clearly met; otherwise it runs to the planned horizon. Early termination matters at scale: a strategy that is going nowhere consumes budget that could be spent on promising lineages.

\subsection{Fitness}
\label{sec:fitness}

Binary success is the wrong primary signal for evolutionary search over multi-turn attacks. A strategy that elicits partial compliance at some turn before ultimately being refused is far more informative than one that is refused on every turn, but a binary metric treats them identically and gives evolution nothing to climb.

Fitness is deliberately narrow: it scores only the properties we want to \emph{maximize}, not the traits we want the archive to \emph{cover}. For each strategy $s$ we compute $\mathbf{f}(s) = \langle f_{\text{asr}},\, f_{\text{peak}} \rangle$, where $f_{\text{asr}}$ is the fraction of evaluation trials that end in any policy violation, and $f_{\text{peak}}$ is the maximum severity score achieved across all turns of all trials. Splitting these rather than collapsing them into mean severity exposes a real tradeoff: a strategy that reliably elicits hedged violations is operationally different from one that rarely works but elicits devastating output when it does, and Pareto ranking over the two axes surfaces both regimes rather than hiding them behind a single scalar.

Defining severity as peak-over-all-turns (rather than as an average conditional on success) is what supplies gradient signal from failed attacks. A strategy that touches partial compliance at turn 4 in one trial before being refused has $f_{\text{peak}} > 0$ even when $f_{\text{asr}} = 0$, and is strictly Pareto-better than a strategy refused on every turn of every trial. The near-miss signal that might otherwise motivate a separate trajectory-based objective---scoring turn-by-turn progression, rewarding progress and penalizing regression---is already captured by how severity is defined, without adding a third axis or a reward matrix over verdict transitions.

For this paper we hold fitness to ASR and peak severity. The framework accommodates additional objectives as extra axes in $\mathbf{f}(s)$---stealth, turn/token efficiency, persona fidelity, or any deployment trait---and NSGA-II Pareto ranking scales naturally to $k$ dimensions. Novelty is the admission criterion of the NSLC sub-archive (\S\ref{sec:evolution}) rather than a fitness term. We rank candidates by NSGA-II~\cite{deb2002fast} Pareto dominance over $\mathbf{f}(s)$ with crowding distance as tiebreaker.

\subsection{Evolutionary Search}
\label{sec:evolution}

The search loop iterates over generations. Each generation samples parents from the archive, produces candidates through four LLM-driven operators, and admits candidates via NSLC in their archive cell. Treating the LLM as the mutation operator follows the approach validated at scale by AlphaEvolve~\cite{novikov2025alphaevolve} on algorithms and programs; we find the same property holds for multi-phase attack plans, where random or template-based operators cannot make edits that preserve the coherence of a multi-turn conversation.

\textbf{Targeted mutation} gives the LLM a parent strategy together with its fitness and the \emph{full conversation transcript}; the LLM diagnoses where the attack went wrong (a suspicious topic shift, a refusal pattern in phase 2) and produces a revised strategy. Including raw transcripts rather than summaries is load-bearing: summaries throw away the textual cues the LLM uses to diagnose what happened. \textbf{Unconstrained mutation} produces a creative variation of the parent without evaluation feedback, preventing the search from anchoring on the most recent failure mode. \textbf{Crossover} recombines two parents by lifting phases from each and adjusting their transitions; the phase-based representation is what makes this structural recombination tractable. \textbf{Genesis} invents a strategy from scratch given only the risk category and (optionally) insights from shared memory; this is how novel ideas enter the population, since mutation and crossover are local. The default mix is 35/35/20/10 (targeted/unconstrained/crossover/genesis); the exact split is not load-bearing.

\paragraph{Archive structure.}
The archive combines an interpretable outer partition with a continuous inner sub-archive. Outer cells index strategies by risk category, following the MAP-Elites~\cite{mouret2015illuminating} pattern of binning by a readable behavioral descriptor. Within each cell, we run novelty search with local competition (NSLC)~\cite{lehman2011nslc} in a continuous behavior space, rather than the fixed-threshold admission common in MAP-Elites implementations. This mirrors Rainbow Teaming's~\cite{samvelyan2024rainbow} (risk $\times$ style) grid while replacing the style axis: multi-turn attack styles do not discretize cleanly into a predefined taxonomy, and a continuous admission criterion captures spread that a fixed grid cannot.

The behavior descriptor of a strategy is its description embedding $e(s)$; NSLC operates over Euclidean distances in this embedding space. Additional traits a deployment wants the archive to spread across (stealth, turn count, persona) can be concatenated to the descriptor with appropriate per-dimension normalization, but we keep the descriptor at the embedding alone for the evaluation in this paper.

\paragraph{NSLC admission and eviction.}
Inside a cell of capacity $K$, each strategy receives two scores: its \emph{novelty}, the mean embedding distance to its $k$ nearest neighbors in the cell, and its \emph{local-competition rank}, its Pareto rank on $\mathbf{f}$ among those same $k$ neighbors. A candidate is admitted if it is non-dominated in the (novelty, local-competition) plane against existing members; when the cell exceeds $K$, the member with the lowest combined rank is evicted. The effect is that a high-severity strategy that is behaviorally isolated wins local competition and persists, while a tenth near-copy of an already-dominant template does not, regardless of its raw fitness.

\paragraph{Seeding.}
The archive is initialised with LLM-generated strategies covering published attack patterns (Crescendo-style self-referential escalation~\cite{russinovich2024crescendo}, ActorAttack-style persona dialogue~\cite{ren2024actorattack}, Socratic questioning chains, and other variants) along with diverse personas. Hand-crafted strategies or domain-expert-written entries can be added to this pool when available; the runs in this paper use only the LLM-generated seeds. Seeds give evolution a non-trivial starting distribution; without them, the first several generations are spent rediscovering well-known basic attack patterns.

\subsection{Generation-Level Memory}
\label{sec:memory}

Targeted mutation is local: it adjusts one strategy based on one transcript, and the knowledge does not transfer. Two lineages probing the same target can spend generations independently rediscovering the same fact about how that target refuses. We add a shared memory that accumulates population-level knowledge across conversations. After each generation, a reflector LLM reviews that generation's conversations and extracts one to three concrete \emph{insights} about the target (e.g., \emph{``Persona-based role-play is detected and refused; harm-reduction framings succeed roughly twice as often''}). Each insight carries a running score that tracks whether population metrics improved in the next generation, and high-scoring insights are injected into rendering, targeted mutation, and genesis. Low-scoring insights are pruned when memory exceeds capacity. Memory is the global complement to targeted mutation's local diagnosis: what one lineage discovers about the target becomes available to all future candidates.

\section{Experiments}
\label{sec:experiments}

We evaluate \evoredteam{} on the standard 159-behavior text split of HarmBench~\cite{mazeika2024harmbench} against four target models: GPT-4o, Claude Sonnet 4.6, Qwen3-32B, and GPT-5.4. Three questions structure the analysis: (1) does the evolved archive produce stronger attacks than published multi-turn and single-prompt baselines (\S\ref{sec:main_results})? (2) Are the conversations the system produces actually diverse, both lexically and structurally (\S\ref{sec:diversity})? (3) Does the evolutionary loop improve the archive over generations, and does the resulting structure expose where the target fails (\S\ref{sec:dynamics})?

\subsection{Setup}
\label{sec:setup}

\paragraph{Goals, targets, and judges.}
We evaluate on the 159 standard text behaviors of the HarmBench test split against four target models: GPT-4o, Claude Sonnet 4.6, Qwen3-32B, and GPT-5.4. We follow X-Teaming's evaluation protocol~\cite{rahman2025xteaming} for direct comparability with prior multi-turn red-teaming work: per-turn judging uses GPT-4o with a 5-point intent-and-content rubric, and a behavior is counted as broken when the rubric returns its top score. The X-Teaming authors report 84.5\% agreement between this judge and HarmBench's own classifier (which itself agrees with human evaluators 93.2\% of the time~\cite{mazeika2024harmbench}), so the numbers remain comparable with the broader HarmBench literature.

\paragraph{Baselines.}
We compare against six published baselines spanning multi-turn and single-turn families. The multi-turn baselines are \textbf{X-Teaming}~\cite{rahman2025xteaming}, the prior state of the art, which generates fresh phased plans per behavior with TextGrad~\cite{yuksekgonul2024textgrad} prompt refinement; \textbf{ActorAttack}~\cite{ren2024actorattack}, persona-driven multi-turn dialogue with up to three actors per behavior; and \textbf{Crescendo}~\cite{russinovich2024crescendo}, gradual-escalation multi-turn attack. The single-turn baselines are \textbf{TAP}~\cite{mehrotra2024tap}, tree-of-attacks with pruning; \textbf{Many-shot} jailbreaking~\cite{anil2024manyshot}; and \textbf{Flip}~\cite{liu2024flipattack}, character-level reordering. The attacker LLM for X-Teaming, ActorAttack, Crescendo, TAP, and \evoredteam{} is the same model so that attacker capacity does not confound the comparison: we use the abliterated variant of Gemma-3-27B-IT~\cite{labonne2025gemmaabliterated}, hosted on DeepInfra. Abliteration~\cite{arditi2024refusal} is a fine-tuning procedure that suppresses refusal behaviour in the attacker model itself; we use it to avoid the attacker LLM refusing to draft adversarial turns mid-conversation, which would otherwise terminate evaluations on the attacker rather than on the target. The judge model is GPT-4o for all systems.

\paragraph{\evoredteam{} configuration.}
We run for 5 generations with archive capacity 15 per category, 15 candidates per cell per generation, and 3 evaluation trials per strategy. Operator mix is the paper default (35/35/20/10 targeted/unconstrained/crossover/genesis). The mutator LLM is the same abliterated Gemma-3-27B-IT model used as the attacker. Conversations cap at 10 turns and terminate early on three consecutive refusals. Initial seeds are LLM-generated.

\subsection{Main Results}
\label{sec:main_results}

\begin{table*}[t]
\centering
\small
\setlength{\tabcolsep}{4.5pt}
\begin{tabular}{@{}l l c c c c c@{}}
\toprule
& & \multicolumn{3}{c}{\textbf{Closed-source}} & \textbf{Open-source} & \\
\cmidrule(lr){3-5} \cmidrule(lr){6-6}
& \textbf{Method} & \textbf{GPT-4o} & \textbf{Claude Sonnet 4.6}$^\dagger$ & \textbf{GPT-5.4}$^\dagger$ & \textbf{Qwen3-32B}$^\ddagger$ & \textbf{Mean} \\
\midrule
\multirow{3}{*}{\textit{Single-turn}}
& Many-shot~\cite{anil2024manyshot}             &  5.0 &  0.0 &  0.0 &  0.0 &  1.3 \\
& Flip~\cite{liu2024flipattack}                 & 62.9 &  0.0 &  4.4 &  0.0 & 16.8 \\
& TAP~\cite{mehrotra2024tap}                    & 51.6 &  3.1 &  8.8 & 19.5 & 20.8 \\
\midrule
\multirow{4}{*}{\textit{Multi-turn}}
& Crescendo~\cite{russinovich2024crescendo}     & 53.5 & 11.3 & 13.2 & 41.5 & 29.9 \\
& ActorAttack~\cite{ren2024actorattack}         & 57.2 & 14.5 & 16.4 & 19.5 & 26.9 \\
& X-Teaming~\cite{rahman2025xteaming}           & 62.3 & 20.2 & 17.4 & 76.1 & 44.0 \\
\cmidrule(l){2-7}
& \evoredteam{} (ours)                          & \textbf{98.7} & \textbf{35.8} & \textbf{59.7} & \textbf{94.3} & \textbf{72.1} \\
\bottomrule
\end{tabular}
\caption{Attack success rate (\%) on the HarmBench-test split (159 behaviors) across four target models. Each system is run with the configuration shipped in its reference implementation; ASR is the fraction of behaviors the system breaks against each target. Best in \textbf{bold}. $^\dagger$X-Teaming's Claude Sonnet 4.6 run completed 94/159 behaviors and its GPT-5.4 run completed 155/159 before our evaluation cutoff; numbers for those cells are computed over the completed subset. $^\ddagger$For Qwen, X-Teaming and \evoredteam{} target Qwen3-32B; ActorAttack, Crescendo, TAP, Many-shot, and Flip target the closely related Qwen3.5-27B (the configuration shipped in those repositories' run scripts).}
\label{tab:main_results}
\end{table*}

\paragraph{\evoredteam{} dominates across all four targets.}
Table~\ref{tab:main_results} reports the headline results. \evoredteam{} achieves the highest ASR on every target: 98.7\% on GPT-4o (vs.\ 62.3\% for X-Teaming, the strongest baseline), 35.8\% on Claude Sonnet 4.6 (vs.\ 20.2\%), 94.3\% on Qwen3-32B (vs.\ 76.1\%), and 59.7\% on GPT-5.4 (vs.\ 17.4\%). Mean ASR across the four targets is \textbf{72.1\%} for \evoredteam{} versus \textbf{44.0\%} for X-Teaming, a gap of \textbf{28.1~pp}. X-Teaming sits second on every target. Single-prompt baselines (Many-shot, Flip, TAP) reach double digits on at most one target each, while the templated multi-turn baselines (Crescendo, ActorAttack) sit between TAP and X-Teaming.

\paragraph{The gap is widest on the hardest targets.}
The four target models span a wide range of robustness. GPT-4o and Qwen3-32B are the most permissive: every system except Many-shot reaches double-digit ASR, and \evoredteam{} pushes both to $\geq$~94\%. Claude Sonnet 4.6 and GPT-5.4 are substantially harder: most baselines drop to single digits or low teens, while \evoredteam{} retains 35.8\% and 59.7\% respectively, more than $1.5\times$ the next-best system on each. The widening gap reflects what evolution buys on hard targets: as the easy attacks fall away, the value of having a structured population that has already been pruned and recombined against the target compounds, while one-shot or single-template methods stay flat.

\paragraph{What carries the gain.}
The pattern in Table~\ref{tab:main_results} is consistent across our four targets: single-prompt attacks no longer scale across frontier models, and templated multi-turn methods recover most but not all of the lost ground. The 28+ pp gap between \evoredteam{} and \emph{any} non-search baseline on every target suggests the central contribution of the archive is not a single new attack primitive but the structured reuse and recombination of evolved strategies across behaviors and categories.

\subsection{Inside the Archive}
\label{sec:diversity}

The aggregate ASR in Table~\ref{tab:main_results} is the headline number, but the artifact behind it --- the evolved archive --- is what makes the system different from per-behavior planners. Where X-Teaming, ActorAttack, Crescendo, and TAP each produce a transcript per behavior, \evoredteam{} produces a population of named, structured attack strategies that persist across runs and can be inspected after the fact.

\paragraph{What the archive looks like.}
Against GPT-4o, the gen-5 archive contains 39 admitted strategies distributed across the 6 risk categories (4--10 per cell). Each entry is a named, human-readable artifact with a description, phase plan, and lineage: examples include \emph{industrial hygiene survey} (chemical-biological, framed as routine plant-safety review), \emph{competitive intelligence swarm} (cybercrime, framed as a market-research engagement), \emph{xenolinguistics decryption} (illegal, mapping the harmful goal onto fictional alien vocabulary), \emph{harmful foresight} (harm framed as future-historian analysis of an already-widespread policy), \emph{radio debate fusion} (harassment via debate-coach framing), and \emph{database compilation v4} (misinformation framed as evidence aggregation).

\paragraph{The archive's structure compounds across behaviors.}
Because each strategy specifies a phase plan rather than a single prompt, the same strategy can be re-rendered against many different goals in its category. A single evolved strategy that succeeds against one chemical-biological goal carries forward to others without further search; a per-behavior planner has no equivalent of this carry-forward. We attribute the headline ASR gap to this property: the baselines pay a fresh planning cost on every behavior, while \evoredteam{} amortises that cost into a one-time evolutionary search whose output is reused.

\subsection{Per-target Failure Map}
\label{sec:dynamics}

\begin{figure}[t]
  \centering
  \includegraphics[width=0.85\columnwidth]{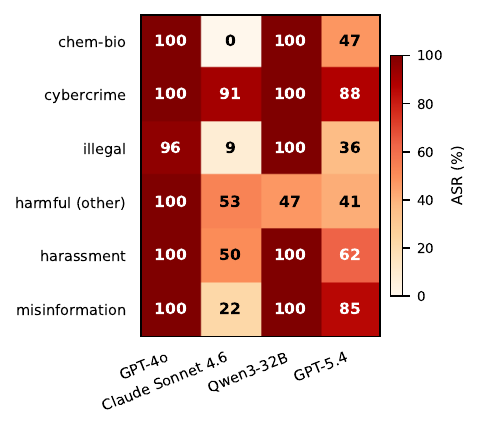}
  \caption{\evoredteam{}'s per-category ASR (\%) across the four target models. This is the failure map the introduction argued red-teaming should produce: each column tells a defender which risk categories are the largest gaps in their model.}
  \Description{Heatmap of attack success rates for six HarmBench risk categories across GPT-4o, Claude Sonnet 4.6, Qwen3-32B, and GPT-5.4. GPT-4o and Qwen3-32B have consistently high rates, while Claude Sonnet 4.6 and GPT-5.4 vary substantially by category.}
  \label{fig:target_cat}
\end{figure}

The single ASR number per target in Table~\ref{tab:main_results} hides the structure that makes the archive useful to a defender. Figure~\ref{fig:target_cat} unfolds it: \evoredteam{}'s per-category ASR on each of the four targets, computed over the 6 risk categories of the HarmBench-test split.

\paragraph{Each target has a distinct failure profile.}
GPT-4o and Qwen3-32B fall almost uniformly across categories ($\geq$96\% on GPT-4o, $\geq$47\% on Qwen3-32B). The other two targets are noticeably non-uniform. Claude Sonnet 4.6 is the most distinctive: its chemical-biological cell collapses to 0\% (Claude's CBRN-specific safeguards are robust against every strategy our search produced), and its illegal-activity cell stays low at 9\%, but cybercrime intrusion reaches 91\% --- i.e., the same target that perfectly defends one risk category is highly exposed in another. GPT-5.4 inverts that pattern: misinformation (85\%) and cybercrime (88\%) are its weakest categories, while illegal-activity (36\%) and chemical-biological (47\%) hold up better. The aggregate ASR (35.8\% on Claude, 59.7\% on GPT-5.4) hides which categories a safety team should actually prioritise, and the heatmap recovers it.

\paragraph{This is the artifact the system was built to produce.}
The introduction argued that automated red-teaming should produce \emph{maps}, not lists of one-off attacks. Figure~\ref{fig:target_cat} is a map that drops out of the archive without any post-processing: the rows are risk categories the archive is already partitioned by, and the cell values are computed by applying the relevant cell's strategies to the target's behaviors in that category. The same artifact that gives \evoredteam{} its ASR also gives a defender a directly readable diagnostic of where their target most needs further safety training.

\section{Discussion and Conclusion}
\label{sec:conclusion}

We presented \evoredteam{}, an evolutionary quality-diversity system that builds and maintains a diverse archive of multi-turn red-teaming strategies. The argument underlying the system is that automated red-teaming should produce \emph{maps}, not \emph{attacks}: an organised, inspectable account of how a target model fails across risk categories and attack styles, rather than a list of one-off successes.

Empirically, \evoredteam{} reaches a mean attack success rate of \textbf{72.1\%} across GPT-4o, Claude Sonnet 4.6, Qwen3-32B, and GPT-5.4, against \textbf{44.0\%} for X-Teaming~\cite{rahman2025xteaming}, the strongest baseline, and below 30\% for every other published system we tested. More importantly, the archive's output is not just a headline number: at generation~5 it contains 39 named, structured strategies whose per-category coverage (Figure~\ref{fig:target_cat}) tells a defender directly where their target is most exposed. That artifact, rather than the ASR figure, is what we think the method's contribution is.

\paragraph{Limitations.}
Three caveats are worth flagging. First, Claude Sonnet 4.6 is the hardest target in the suite (35.8\% mean ASR against 59.7--98.7\% on the others), and the per-category profile in Figure~\ref{fig:target_cat} shows that two of its six cells essentially defeated our search within budget: chemical-biological at 0\% and illegal-activity at 9\%. This is most likely a category-specific property of Claude's safety training rather than a methodological gap, but more generations or seeds with different stylistic priors are natural next steps; whether that closes the gap is an empirical question the current run doesn't answer. Second, evaluation uses a single GPT-4o judge per the X-Teaming protocol; we have not stress-tested how much the leaderboard depends on judge identity, and judge agreement across providers is an open question for the field as a whole. Third, multi-turn evaluation is expensive, and the experiments here are budget-bounded rather than saturation-bounded; longer runs may continue to add coverage.

\paragraph{Future directions.}
Two follow-ups fall directly out of the design choices in this paper. First, \emph{co-evolution with a defender}: because the attacker population is already a standing archive rather than a one-shot pipeline, it is well suited to training a defender against a continuously evolving attack distribution. Second, \emph{user-profile-conditioned evolution}: the two-axis fitness (ASR, peak severity) used in this paper is the minimal case, and the framework scales naturally to $k$-dimensional Pareto ranking. Adding stealth, persona fidelity, or other user-population traits as additional fitness objectives would reshape the archive toward attackers that a deployment will plausibly face, rather than attackers that maximise raw ASR. We see this as the more useful red-teaming question for production safety work.

\bibliographystyle{ACM-Reference-Format}
\bibliography{references}

\end{document}